# SATZ - An Adaptive Sentence Segmentation System

*David D. Palmer*

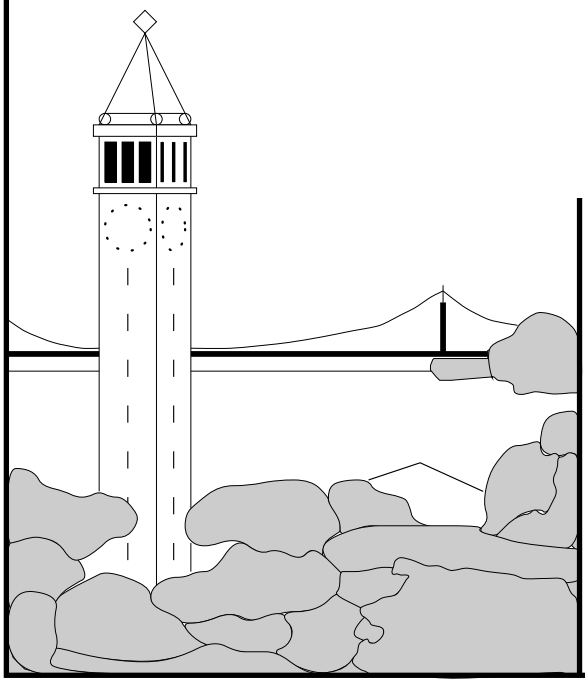

# SATZ - An Adaptive Sentence Segmentation System
## David D. Palmer

## Abstract

The segmentation of a text into sentences is a necessary prerequisite for many natural language processing tasks, including part-of-speech tagging and sentence alignment. This is a non-trivial task, however, since end-of-sentence punctuation marks are ambiguous. A period, for example, can denote a decimal point, an abbreviation, the end of a sentence, or even an abbreviation at the end of a sentence. To disambiguate punctuation marks most systems use brittle, special-purpose regular expression grammars and exception rules. Such approaches are usually limited to the text genre for which they were developed and cannot be easily adapted to new text types. They can also not be easily adapted to other natural languages.

As an alternative, I present an efficient, trainable algorithm that can be easily adapted to new text genres and some range of natural languages. The algorithm uses a lexicon with part-of-speech probabilities and a feed-forward neural network for rapid training. The method described requires minimal storage overhead and a very small amount of training data. The algorithm overcomes the limitations of existing methods and produces a very high accuracy.

The results presented demonstrate the successful implementation of the algorithm on a 27,294 sentence English corpus. Training time was less than one minute on a workstation and the method correctly labeled over 98.5% of the sentence boundaries. The method was also successful in labeling texts containing no capital letters. The system has been successfully adapted to German and French. The training times were similarly low and the resulting accuracy exceeded 99%.

# Contents

# 1  Introduction

## 1.1  The Problem

The sentence is an important unit in many natural language processing tasks.[1] For example, the alignment of sentences in parallel multi-lingual corpora requires first that the individual sentence boundaries be clearly labeled (Gale and Church, 1993), (Kay and Röscheinsen, 1993). Most part-of-speech taggers also require the disambiguation of sentence boundaries[2] in the input text (Church, 1988),(Cutting et al., 1991). This is usually accomplished by inserting a unique character sequence at the end of each sentence, such that the NLP tools analyzing the text can easily recognize the individual sentences.

Segmenting a text into sentences is a non-trivial task, however, since all end-of-sentence punctuation marks[3] are ambiguous. A period, for example, can denote a decimal point, an abbreviation, the end of a sentence, or even an abbreviation at the end of a sentence. An exclamation point and a question mark can occur within quotation marks or parentheses, as well as at the end of a sentence. The ambiguity of these punctuation marks is illustrated in the following difficult cases:

(1) *The group included Dr. J.M. Freeman and T. Boone Pickens Jr.*

(2) *"This issue crosses party lines and crosses philosophical lines!"  said Rep. John Rowland (R., Conn.).*

The existence of punctuation in grammatical subsentences suggests the possibility of a further decomposition of the sentence boundary problem into types of sentence boundaries, one of which would be "embedded sentence boundary." Such a distinction might be useful for certain applications which analyze the grammatical structure of the sentence. However, in this work I will address the less-specific problem of determining sentence boundaries between sentences.

---

[1]Much of the work contained in this report has been reported in a similar form in (Palmer and Hearst, 1994) and portions of this work were done in collaboration with Marti Hearst of Xerox PARC.

[2]The terms *sentence segmentation*, *sentence boundary disambiguation* and *sentence boundary labeling* are interchangeable.

[3]In this report, I will consider only the period, the exclamation point and the question mark to be possible "end-of-sentence punctuation marks", and all references to "punctuation marks" will refer to these three. Although the colon, the semicolon, and conceivably the comma can also delimit grammatical sentences, their usage is beyond the scope of this work.

In examples (1) and (2), the word immediately preceding and the word immediately following a punctuation mark provide important information about its role in the sentence. However, more context may be necessary, such as when punctuation occurs in a subsentence within quotation marks or parentheses, as seen in example (2), or when an abbreviation appears at the end of a sentence, as seen in (3a-b):

(3a)  *It was due Friday by 5 p.m. Saturday would be too late.*

(3b)  *She has an appointment at 5 p.m. Saturday to get her car fixed.*

Section 4.2.1 contains a discussion of methods of representing context.

## 1.2    Baseline

When evaluating a sentence segmentation algorithm, comparison with the *baseline algorithm* is an important measure of the success of the algorithm. A baseline algorithm in this case is simply a very naive algorithm which would label each punctuation mark as a sentence boundary. Such a baseline algorithm would have an accuracy equal to the *lower bound* of the text, the percentage of possible sentence-ending punctuation marks in the text which indeed denote sentence boundaries. A good sentence segmentation algorithm will thus have an accuracy much greater than the lower bound.

Since the use of abbreviations in a text depends on the particular text and text genre, the number of ambiguous punctuation marks, and therefore the performance of the baseline algorithm, will vary dramatically depending on text genre, and even within a single text genre. For example, Liberman and Church (1992) report that the Wall Street Journal corpus contains 14,153 periods per million tokens, whereas, in the Tagged Brown corpus (Francis and Kucera, 1982), the figure is only 10,910 periods per million tokens. They also report that 47% of the periods in the WSJ corpus denote abbreviations (lower bound 53%), compared to only 10% in the Brown corpus (lower bound 90%)(Riley, 1989). In contrast, Müller (1980) reports lower bound statistics ranging from 54.7% to 92.8% within the same corpus of scientific abstracts. Such a range of lower bound figures might suggest the need for a robust approach that can adapt rapidly to different text requirements.

# 2 Previous Approaches

Although sentence boundary disambiguation is an essential preprocessing step of many natural language processing systems, it is a topic rarely addressed in the literature. Consequently, there are few published references. There are also few public domain systems for performing the segmentation task, and most current systems are specifically tailored to the particular corpus analyzed and are not designed for general use.

## 2.1 Regular Expressions and Heuristic Rules

The method currently widely used for determining sentence boundaries is a regular grammar, usually with limited lookahead. In the simplest implementation of this method, the grammar rules attempt to find patterns of characters, such as "period-space-capital letter" which usually occur at the end of a sentence. More robust implementations consider the entire word preceding and following the punctuation mark and include extensive word lists and exception lists to attempt to recognize abbreviations and proper nouns. There are several examples of rule-based and heuristic systems for which performance numbers are available.

Christiane Hoffmann(1994) used a regular expression approach to classify punctuation marks in a corpus of the German newspaper *die tageszeitung* with a lower bound of 92%. She used the UNIX tool lex (Lesk and Schmidt, 1975) and a large abbreviation list to classify occurrences of periods according to their likely function in the text. Tested on 2827 periods from the corpus, her method correctly classified over 98% of the sentence boundaries. The method was developed specifically for the *tageszeitung* corpus, and Hoffmann reports that success in applying her method to other corpora would be dependent on the quality of the available abbreviation lists.

Gabriele Schicht[4], over the course of four months, developed a method for segmenting sentences in a corpus of the German newspaper *die Süddeutsche*

---

[4]At the University of Munich, Germany.

*Zeitung.* The method uses a program written in the text manipulation language perl (Wall and Schwartz, 1991) to analyze a context consisting of the word immediately preceding and the word immediately following each punctuation mark. In the case of a period following a number the method considers more context, one word before the number and one word after the period. More context is also considered when attempting to recognize abbreviations containing several blank spaces, such as "v. i. S. d. P." (verantwortlich im Sinne des Presserechts). Using a Next workstation, the method requires 30 minutes to classify 15,000-20,000 cases, as each word (in the limited context) must be looked up in a 500,000 word lexicon. Schicht reports over 99% accuracy using the method.[5]

Mark Wasson and colleagues[6] invested 9 staff months developing a system that recognizes special tokens (e.g., non-dictionary terms such as proper names, legal statute citations, etc.) as well as sentence boundaries. From this, Wasson built a stand-alone boundary recognizer in the form of a grammar converted into finite automata with 1419 states and 18002 transitions (excluding the lexicon). The resulting system, when tested on 20 megabytes of news and case law text achieved an accuracy of 99.7% at speeds of 80,000 characters per CPU second on a mainframe computer. When tested against upper-case legal text the algorithm still performed very well, achieving accuracies of 99.71% and 98.24% on test data of 5305 and 9396 periods, respectively. It is not likely, however, that the results would be this strong on lower-case data.[7]

Although the regular grammar approach can be successful, it requires a large manual effort to compile the individual rules used to recognize the sentence boundaries. Such efforts are usually developed specifically for a text corpus (Liberman and Church, 1992), (Hoffmann, 1994), and would probably not be portable to other text genres. Because of their reliance on special language-specific word lists, they are not portable to other natural languages without repeating the effort of compiling extensive lists and rewriting rules. In addition, heuristic approaches depend on having a well-behaved corpus with regular punctuation and few extraneous characters, and they would probably not be very successful with texts obtained via optical character recognition (OCR).

---

[5] All information about this system is courtesy of a personal communication with Gabriele Schicht.

[6] At Mead Data Central.

[7] All information about Mead's system is courtesy of a personal communication between Marti Hearst and Mark Wasson.

## 2.2 Regression Trees

Riley (1989) describes an approach that uses regression trees (Breiman et al., 1984) to classify sentence boundaries according to the following features:

Probability[word preceding "." occurs at end of sentence]

Probability[word following "." occurs at beginning of sentence]

Length of word preceding "."

Length of word after "."

Case of word preceding ".": Upper, Lower, Cap, Numbers

Case of word following ".": Upper, Lower Cap, Numbers

Punctuation after "." (if any)

Abbreviation class of words with "."

The method uses information about one word of context on either side of the punctuation mark and thus must record, for every word in the lexicon, the probability that it occurs next to a sentence boundary. Probabilities were compiled from 25 million words of pre-labeled training data from a corpus of AP newswire. The results were tested on the Brown corpus achieving an accuracy of 99.8%.[8]

## 2.3 Word endings and word lists

Müller (1980) provides an exhaustive analysis of sentence boundary disambiguation as it relates to lexical endings and the identification of abbreviations and words surrounding a punctuation mark, focusing on text written in English. This approach makes multiple passes through the data to find recognizable suffixes and thereby filters out words which aren't likely to be abbreviations. The morphological analysis makes it possible to identify words which are not otherwise present in the extensive word lists used to identify abbreviations. Accuracy rates of 95-98% are reported for this method tested on over 75,000 scientific abstracts, with a lower bound ranging from 54.7% to 92.8%.

---

[8]Time for training was not reported, nor was the amount of the Brown corpus against which testing was performed; it is assumed the entire Brown corpus was used.

## 2.4   Feed-forward Neural Network

Humphrey and Zhou (1989) report using a feed-forward neural network to disambiguate periods, and achieve an accuracy averaging 93%. They use a regular grammar to tokenize the text before training the neural nets, but no further details of their approach are available.[9]

# 3   System Desiderata

Each of the approaches described above has disadvantages to overcome. A successful sentence-boundary disambiguation algorithm should have the following characteristics:

- The approach should be robust, and should not require a hand-built grammar or specialized rules that depend heavily on capitalization, multiple spaces between sentences, etc. Thus, the approach should adapt easily to new text genres and some new languages.

- The approach should train quickly on a small training set and should not require excessive storage overhead.

- The approach's results should be very accurate and it should be efficient enough that it does not noticeably slow down text preprocessing.

- The approach should be able to specify "no opinion" on cases that are too difficult to disambiguate, rather than making under-informed guesses.

---

[9]Accuracy results were obtained courtesy of a personal communication between Marti Hearst and Joe Zhou.

In the following sections I present an approach that meets each of these criteria, produces a very low error rate, and behaves more robustly than solutions that require manually designed rules.

# 4  The SATZ System

This section describes the structure of my adaptive sentence segmentation system, known as SATZ[10]. My approach in the SATZ system is to represent the context surrounding a punctuation mark as a series of vectors of probabilities. The probabilities used for each word in the context are the prior part-of-speech probabilities obtained from a lexicon containing part-of-speech frequency data. The context vectors, or "descriptor arrays," are used as input to a neural network trained to disambiguate sentence boundaries. The output of the neural network is then used to determine the role of the punctuation mark in the sentence. The architecture of the system is shown in Figure 1, and the following sections describe the individual stages in the process.

## 4.1  Tokenizer

The first stage of the process is lexical analysis, which breaks the input text (a stream of characters) into tokens. The SATZ tokenizer is implemented using the UNIX tool lex (Lesk and Schmidt, 1975) and is a slightly-modified version of the tokenizer from the PARTS part-of-speech tagger (Church, 1988). The tokens returned by the lex program can be a sequence of alphabetic characters (i.e. words), a sequence of digits[11], or a single non-alphanumeric character such as a period or quotation mark.

## 4.2  Part-of-speech Lookup

### 4.2.1  Representing Context

The context surrounding a punctuation mark can be represented in various ways. The simplest and most straightforward is to use the individual words

---

[10] "Satz" is the German word for "sentence."

[11] Numbers containing periods acting as decimal points are considered a single token. This eliminates one possible ambiguity of the period at the lexical analysis stage.

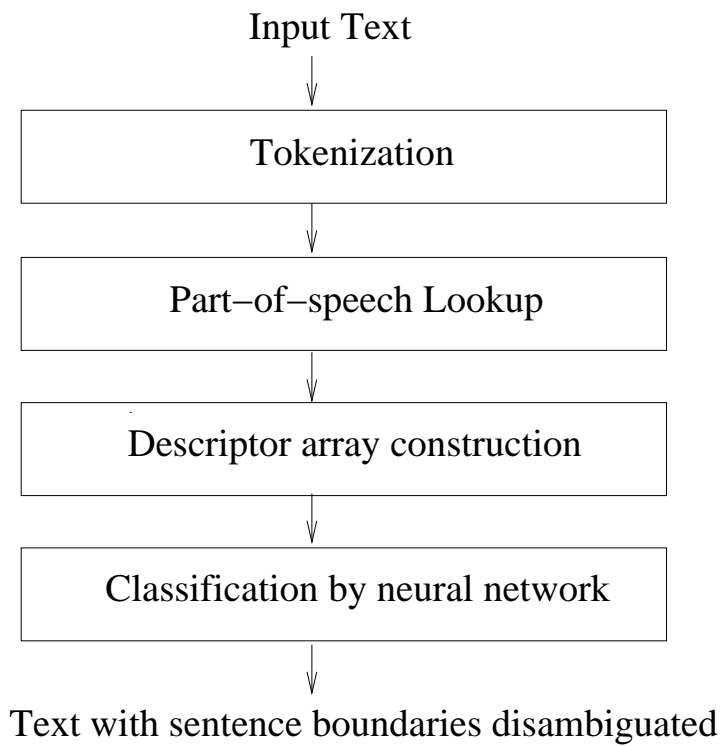

Figure 1: SATZ Architecture

preceding and following the punctuation mark, as in this example using three words of context on either side of the punctuation mark:

*at the plant. He had thought*

For each word in the language, we would then determine how likely it is to come at the end or beginning of a sentence. However, compiling these figures for each word in a language is very time-consuming and requires large amounts of storage, and it is unlikely that such information will be useful to later stages of processing.

As an alternative, the context could be approximated by using a single part-of-speech for each word. The above context would then be represented by the following part-of-speech sequence:

*preposition article noun*

*pronoun verb verb*

Requiring a single part-of-speech for each word presents a processing circularity: because most part-of-speech taggers require predetermined sentence boundaries, sentence labeling must be done before tagging. But if sentence labeling is done before tagging, no part-of-speech assignments are available for the boundary-determination algorithm.

To avoid this processing circularity and avoid the need for a single part-of-speech for each word, the context can be further approximated by *the prior probabilities* of all parts-of-speech for that word. Each word in the context would thus be represented by a series of possible parts-of-speech, as well as the probability that the word occurs as each part-of-speech. Continuing the example, the context becomes:

*preposition(1.0) article(1.0) noun(0.8)/verb(0.2)*

*pronoun(1.0) verb(1.0) noun(0.1)/verb(0.9)*

This denotes that "at" and "the" have a 1.0 probability of occurring as a preposition and article respectively, "plant" has a 0.8 probability of occurring as a noun and a 0.2 probability of occurring as a verb, and so on. These probabilities are based on occurrences of the words in a pre-tagged corpus, and are therefore corpus dependent. Such part-of-speech information is often used by a part-of-speech tagger and would thus be readily available and would not require excessive storage overhead. For these reasons I chose to approximate the context in my system by using the prior part-of-speech probabilities.

### 4.2.2   The Lexicon

An important component of the SATZ system is the lexicon containing part-of-speech frequency data from which the probabilities are calculated. Words in the lexicon are followed by a series of part-of-speech tags and associated frequencies, representing the possible parts-of-speech for that word and the frequency with which the word occurs as each part-of-speech. The lexical lookup stage of the SATZ system finds a word in the lexicon (if it is present) and returns the possible parts-of-speech. For the English word "well," for example, the lookup module might return the tags

JJ/15 NN/18 QL/68 RB/634 UH/22 VB/5

indicating that, in the corpus on which the lexicon is based[12], the word "well" occurred 15 times as an adjective, 18 as a singular noun, 68 as a qualifier, 634 as an adverb, 22 as an interjection, and 5 as a singular verb.

### 4.2.3 Heuristics for Unknown Words

If a word is not present in the lexicon, the system contains a set of heuristics which attempt to assign the most reasonable parts-of-speech to the word. A summary of these heuristics follows.

- Unknown tokens containing a digit (0-9) are assumed to be numbers.

- Any token beginning with a period, exclamation point, or question mark is assigned a "possible end-of-sentence punctuation" tag. This catches common sequences like "?!"

- Common morphological endings are recognized and the appropriate part-of-speech is assigned to the entire word.

- Words containing a hyphen are assigned a series of tags and frequencies denoting "unknown hyphenated word."

- Words containing an internal period are assumed to be abbreviations.

- Capitalized words are not always proper nouns, even when it appears somewhere other than in a sentence's initial position (e.g., the word "American" is often used as an adjective). Those words not present in the lexicon are assigned a certain probability (0.9 for English) of being a proper noun.

---

[12]In this example, the frequencies are derived from the Brown corpus (Francis and Kucera, 1982).

- Capitalized words appearing in the lexicon but not registered as proper nouns can nevertheless still be proper nouns. In addition to the part-of-speech frequencies present in the lexicon, these words are assigned a certain probability of being a proper noun (0.5 for English).

- As a last resort, the word is assigned a series of possible tags with a uniform frequency distribution.

These heuristics can be easily modified and adapted to the specific needs of a new language. For example, the probability of a capitalized word being a proper noun is higher in English than in German, where all nouns are also capitalized.

## 4.3 Descriptor Array Construction

For each token in the input text, we need to construct a vector of probabilities to numerically describe the token. This vector is known as a *descriptor array*. The lexicon may contain as many as 70 or 80 very specific parts-of-speech, which we first need to map into more general categories. For example, the tags for *present tense verb*, *past participle*, and *modal verb* all map into the more general "verb" category. The parts-of-speech returned by the lookup module are thus mapped into the 18 general categories given in Figure 2, and the frequencies for each category are summed. The 18 category frequencies for the word are then converted to probabilities by dividing the frequencies for each. In addition to these 18 probabilities, the descriptor array also contains two additional flags that indicate if the word begins with a capital letter and if it follows a punctuation mark, for a total of 20 items in each descriptor array.

## 4.4 Classification by Neural Network

The descriptor arrays representing the tokens in the context are used as the input to a fully-connected feed-forward neural network, shown in Figure 3.

### 4.4.1 Network architecture

The network accepts as input $k * 20$ input units, where $k$ is the number of words of context surrounding an instance of an end-of-sentence punctuation mark (referred to in this report as "k-context"), and 20 is the number of elements in the descriptor array described in Section 4.3. The input layer is fully connected to a hidden layer consisting of $j$ hidden units with a sigmoidal squashing activation function. The hidden units in turn feed into one output unit which indicates the results of the function.[13]

| | |
|---|---|
| noun | verb |
| article | modifier |
| conjunction | pronoun |
| preposition | proper noun |
| number | comma or semicolon |
| left parentheses | right parentheses |
| non-punctuation character | possessive |
| colon or dash | abbreviation |
| sentence-ending punctuation | others |

Figure 2: Elements of the descriptor array assigned to each incoming token.

The output of the network, a single value between 0 and 1, represents the strength of the evidence that a punctuation mark occurring in its context is indeed the end of the sentence. I define two adjustable sensitivity thresholds, $t_0$ and $t_1$, which are used to classify the results of the disambiguation. If the output is less than $t_0$, the punctuation mark is not a sentence boundary; if the output is greater than or equal to $t_1$, it is a sentence boundary. Outputs which fall between the thresholds cannot be disambiguated by the network and are marked accordingly, so they can be treated specially in later processing. When $t_0 = t_1$, no punctuation mark is left ambiguous. Section 5.4 describes experiments which vary the sensitivity thresholds.

To disambiguate a punctuation mark in a k-context, a window of $k + 1$ tokens and their descriptor arrays is maintained as the input text is read. The first $k/2$ and final $k/2$ tokens of this sequence represent the context in which the middle token appears. If the middle token is a potential end-of-sentence punctuation mark, the descriptor arrays for the context tokens are input to the network and the output result indicates the appropriate label, subject to the thresholds $t_0$ and $t_1$.

---

[13]This network can be thought of roughly as a Time-Delay Neural Network (TDNN) (Hertz et al., 1991), since it accepts a sequence of inputs and is sensitive to positional information within the sequence. However, since the input information is not really shifted with each time step, but rather only presented to the neural net when a punctuation mark is in the center of the input stream, this is not technically a TDNN.

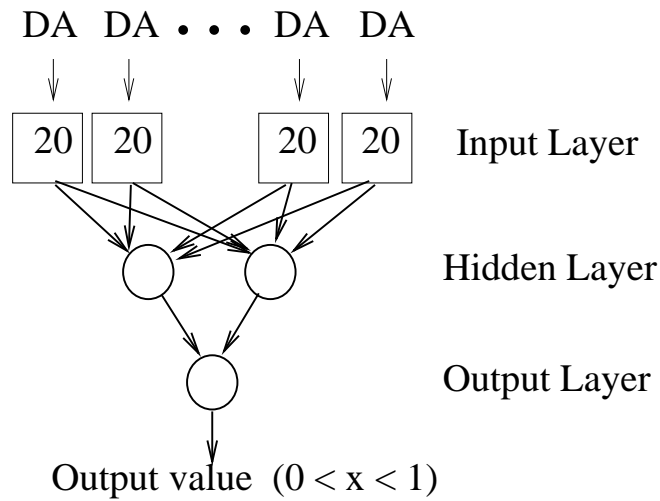

Figure 3: Neural Network Architecture (DA = descriptor array of 20 items)

### 4.4.2 Training

Training data consist of two texts in which all boundaries are already labeled. The first text, the *training text*, contains between 250 and 500 *test cases*, where a test case is an ambiguous punctuation mark. The weights of the neural network are trained on the training text using the standard backpropagation algorithm (Hertz et al., 1991). The second text used in training is the *cross-validation text* (Bourland and Morgan, 1994) and contains between 125 and 250 test cases separate from the training text. Training of the weights is not performed on this text; the cross-validation text is instead used to increase the generalization of the training, such that when the total training error over the cross-validation text reaches a minimum, training is halted. Testing is then performed on texts independent of the training and cross-validation texts. All training times reported in this report were obtained on a Hewlett Packard 9000/750 Workstation, unless otherwise noted.

## 4.5  Implementation

I implemented the SATZ system as a series of C modules and UNIX shell scripts. The software is available via anonymous ftp to cs-tr.CS.Berkeley.EDU in the directory pub/cstr/ as the compressed tar file satz.tar.Z. Appendix A contains the README file I wrote to explain the structure of the software and to assist users in adapting it for their own purposes.

# 5  Experiments and Results (English)

I tested the SATZ system for the English language using texts from the Wall Street Journal portion of the ACL/DCI collection (Church and Liberman, 1991). I first constructed a training text[14] of 573 test cases and a cross-validation text of 258 test cases from the WSJ corpus. I then constructed a separate test text consisting of 27,294 test cases, with a lower bound of 75.0%. The lexicon and thus the frequency counts used to calculate the descriptor arrays were taken from the PARTS tagger (Church, 1988), which derived the counts from the Brown corpus (Francis and Kucera, 1982).

## 5.1  Context Size

In order to determine how much context is necessary to accurately segment sentences in a text, I varied the size of the context and obtained the results in table 1. The *Training Error* is the least mean squares (Hertz et al., 1991) error[15] (one-half the sum of the squares of all the errors) for all 573 items in the training set. The *Cross Error* is the equivalent value for the cross-validation set. These two error figures give an indication of how well the network learned the training data before stopping. From these data I concluded that a 6-token context, 3 preceding the punctuation mark and 3 following, produces the best results.

---

[14]Note that "constructing" a training, cross-validation, or test text simply involves manually inserting a unique character sequence at the end of each sentence.

[15]The *error* of a particular item is the difference between the desired output and the actual output of the neural net.

| Context | Training | Training | Cross | Testing | Testing |
|---|---|---|---|---|---|
| Size | Epochs | Error | Error | Errors | Error (%) |
| 4-context | 1731 | 1.52 | 2.36 | 1424 | 5.22% |
| 6-context | 218 | 0.75 | 2.01 | 409 | 1.50% |
| 8-context | 831 | 0.043 | 1.88 | 877 | 3.21% |

Table 1: Results of comparing context sizes.

| # Hidden | Training | Training | Cross | Testing | Testing |
|---|---|---|---|---|---|
| Units | Epochs | Error | Error | Errors | Error (%) |
| 1 | 623 | 1.05 | 1.61 | 721 | 2.64% |
| 2 | 216 | 1.08 | 2.18 | 409 | 1.50% |
| 3 | 239 | 0.39 | 2.27 | 435 | 1.59% |
| 4 | 350 | 0.27 | 1.42 | 1343 | 4.92% |

Table 2: Results of comparing hidden layer sizes (6-context).

## 5.2   Hidden Units

To determine the size of the hidden layer in the neural network which produced the highest output accuracy, I experimented with various hidden layer sizes and obtained the results in table 2. From these data I concluded that the best accuracy in this case is possible using a neural network with two nodes in its hidden layer.

## 5.3   Sources of Errors

As described in Sections 5.1 and 5.2, the best results were obtained with a context size of 6 tokens and a hidden layer with 2 units. This configuration produced a total of 409 errors out of 27,294 test cases, for an accuracy of 98.5%. These errors fall into two major categories: (i)"false positive", i.e., a punctuation mark the method erroneously labeled as a sentence boundary, and (ii) "false negative", i.e., an actual sentence boundary which the method did not label as such. Table 3 contains a summary of these errors.

These errors can be decomposed into the following groups:

37.6%  false positive at an abbreviation within a title or name, usually because the word following the period exists in the lexicon with other parts-of-speech (*Mr. Gray, Col. North, Mr. Major, Dr. Carpenter, Mr. Sharp*).

| | |
|---|---|
| 224 (54.8%) false positives | |
| 185 (45.2%) false negatives | |
| 409 total errors out of 27,294 items | |

Table 3: Results of testing on 27,294 mixed-case items; $t_0 = t_1 = 0.5$, 6-context, 2 hidden units.

22.5% false negative due to an abbreviation at the end of a sentence, most frequently *Inc., Co., Corp.,* or *U.S.*, which all occur within sentences as well.

11.0% false positive or negative due to a sequence of characters including a period and quotation marks, as this sequence can occur both within and at the end of sentences.

9.2% false negative resulting from an abbreviation followed by quotation marks; related to the previous two types.

9.8% false positive or false negative resulting from presence of ellipsis (...), which can occur at the end of or within a sentence.

9.9% miscellaneous errors, including extraneous characters (dashes, asterisks, etc.), ungrammatical sentences, misspellings, and parenthetical sentences.

The first two items indicate that the system is having difficulty recognizing the function of abbreviations. I attempted to counter this by dividing the abbreviations in the lexicon into two distinct categories, "title abbreviations" such as *Mr.* and *Dr.* which almost never occur at the end of a sentence, and all other abbreviations. This new classification, however, significantly increased the training time and eliminated only 12 of the 409 errors (2.9%).

The third and fourth items demonstrate the difficulty of distinguishing subsentences within a sentence. This problem may be addressed by creating a new classification for punctuation marks, the "embedded end-of-sentence," as discussed in section 1.1. The fifth class of error may similarly be addressed by creating a new classification for ellipses, and then attempting to determine the role of the ellipses independent of the sentence boundaries.

| Lower Thresh | Upper Thresh | False Pos | False Neg | Not Labeled | Were Correct | % Not Labeled | Testing Error (%) |
|---|---|---|---|---|---|---|---|
| 0.5 | 0.5 | 209 | 200 | 0 | 0 | 0.0 | 1.50 |
| 0.4 | 0.6 | 173 | 174 | 145 | 83 | 0.50 | 1.27 |
| 0.3 | 0.7 | 140 | 148 | 326 | 205 | 1.20 | 1.06 |
| 0.2 | 0.8 | 111 | 133 | 541 | 376 | 1.98 | 0.89 |
| 0.1 | 0.9 | 79 | 94 | 1021 | 785 | 3.74 | 0.63 |

Table 4: Results of varying the sensitivity thresholds (27,294 test cases, $t_0 = t_1 = 0.5$, 6-context, 2 hidden units).

## 5.4 Thresholds

As described in Section 4.4.1, the output of the neural network is used to determine the function of a punctuation mark based on its value relative to two sensitivity thresholds, with outputs that fall between the thresholds denoting that the function of the punctuation mark is still ambiguous. These are shown in the "Not Labeled" column of table 4, which gives the results of a systematic experiment with the sensitivity thresholds. As the thresholds were moved from the initial values of 0.5 and 0.5, certain items which had been classified as "False Pos" or "False Neg" fell between the thresholds and became "Not Labeled." At the same time, however, items which had been correctly labeled also fell between the thresholds, and these are shown in the "Were Correct" column[16]. There is thus a tradeoff: decreasing the error percentage by adjusting the thresholds also decreases the percentage of cases correctly labeled and increases the percentage of items left ambiguous.

## 5.5 Single-case texts

A major advantage of the SATZ approach to sentence segmentation is its robustness. In contrast to many existing systems which depend on brittle parameters such as capitalization or spacing, SATZ is able to adapt to texts which are not well-formed, such as single-case texts. The two descriptor array flags for capitalization, discussed in section 4.3, allow the system to include capitalization information when it is available. When this information is not available, the system is nevertheless able to adapt and produce a high accuracy. To demonstrate this robustness, I converted the training, cross-validation, and test texts used in previous testing to a lower-case-only format, with no capital letters. After retraining the neural network with the lower-case-only texts, the SATZ system was able to correctly disambiguate 96.2% of the sentence boundaries. After converting the texts to an upper-case-only format, with all capital letters, and retraining the network on the texts in this format, the system was able to correctly label 97.4%[17]. These results are summarized in table 5.

---

[16] Note that the number of items in the "Were Correct" column is a subset of those in the "Not Labeled" column.

| Text | Training | Training | Training | Cross | Testing |
|------|----------|----------|----------|-------|---------|
| Type | Time (sec) | Epochs | Error | Error | Error (%) |
| Lower-case | 30 | 66 | 7.65 | 2.64 | 3.8% |
| Upper-case | 40 | 192 | 6.33 | 3.46 | 2.6% |

Table 5: Results on single-case texts (27,294 test cases, $t_0 = t_1 = 0.5$, 6-context, 2 hidden units).

## 5.6 Lexicon size

The lexicon with which I obtained the results of the previous sections was the complete lexicon (over 30,000 words) from the PARTS tagger. Such a large lexicon with part-of-speech frequency data is not always available, so it is important to understand the impact a more limited lexicon would have on the accuracy of SATZ. I altered the size of the English lexicon used in training and testing[18], and obtained the results in table 6. These data demonstrate that a larger lexicon provides faster training and a higher accuracy, although the performance with the smaller lexica was still almost as accurate as before.

---

[17]The difference in results with upper-case and lower-case formats can probably be attributed to the capitalization flags in the descriptor arrays.

[18]The abbreviations in the lexicon remained unchanged. Altering the list of abbreviations might be an interesting future experiment.

| Words in Lexicon | Training Epochs | Training Error | Cross Error | Testing Errors | Testing Error (%) |
|---|---|---|---|---|---|
| 30,000 | 218 | 0.75 | 2.01 | 411 | 1.50% |
| 5,000 | 372 | 0.50 | 1.75 | 483 | 1.75% |
| 3,000 | 1056 | 0.05 | 1.30 | 551 | 2.00% |

Table 6: Results of comparing lexicon size (27,294 test cases, $t_0 = t_1 = 0.5$, 6-context, 2 hidden units).

# 6 Adaptation to Some Other Languages

Since the disambiguation component of the sentence segmentation algorithm, the neural network, is language independent, the SATZ system can be easily adapted to some natural languages other than English. Adaptation to other languages involves setting a few language-specific parameters, and obtaining (or building) a small lexicon containing the necessary part-of-speech data. I successfully adapted the SATZ system to German and French, and the results are described below.

## 6.1 German

The German lexicon was built from a series of public-domain word lists obtained from the Consortium for Lexical Research. The lists of German adjectives, verbs, prepositions, articles, and abbreviations were converted to the appropriate format described in Section 4.2.2. In the resulting lexicon of 18,000 words, each word was assigned only the parts-of-speech for the lists from which it came, with a frequency of 1 for each part-of-speech. The lexicon contained 156 German abbreviations. The part-of-speech tags used were identical to those from the English lexicon, and the descriptor array mapping remained also unchanged. This lexicon was used in testing with two separate corpora. The total time required to adapt SATZ to German, including building the lexicon and constructing training texts, was less than one day.

### 6.1.1 German News Corpus

The German News Corpus was constructed from a series of public-domain German articles distributed internationally by the University of Ulm. It contained over 12,000 test cases from the months July-October 1994, with a lower bound of 96.7%. I constructed a training text of 268 cases from the corpus, as well as a cross-validation text of 150 cases. The training was completed in 70 seconds and resulted in a rate of 99.0% correctly labeled sentence boundaries in the corpus of 12,000 cases. Repeating the training and testing with a lower-case-only format gave an accuracy rate of 99.3%. This higher accuracy for the lower-case text might be a result of the German capitalization rules, in which all nouns, not just proper nouns, are capitalized. I performed the training (both mixed-case and lower-case) without altering the heuristics described in section 4.2.3. Fine-tuning the probabilities for unknown capitalized words in German may increase the mixed-case accuracy.

### 6.1.2 Süddeutsche Zeitung Corpus

The Süddeutsche Zeitung Corpus compiled at the University of Munich consists of several megabytes of online texts from the German newspaper.[19] I constructed a training text of over 500 items from the Süddeutsche Zeitung Corpus, and a cross-validation text of over 250 items. Training was performed in less than 5 minutes on a Next workstation[20]. When tested on the September 1994 portion of the SZ corpus containing approximately 20,000 items, the SATZ system produced an accuracy comparable to[21] those obtained with Schicht's method described in Section 2.1.

## 6.2 French

The French lexicon was compiled from the part-of-speech data obtained by running the PARC part-of-speech tagger (Cutting et al., 1991) on a portion of the Canadian Hansards corpus.[22] The lexicon consisted of less than 1000 words assigned parts-of-speech by the tagger, including 20 French abbreviations appended to the 200 English abbreviations available from the English lexicon. The part-of-speech tags in the lexicon were different from those used in the English implementation, so the descriptor array mapping had to be adjusted accordingly. Adapting SATZ to French was accomplished in 2 days.

---

[19]All my work with the Süddeutsche Zeitung Corpus was performed in collaboration with Prof. Franz Guenthner and Gabriele Schicht of the Centrum für Informations- und Sprachverarbeitung at the University of Munich.

[20]The Next workstation is significantly slower than the Hewlett Packard workstation used in other tests, which accounts for the slower training time.

[21]Due to the large size of the corpus, it was impossible to obtain an exact accuracy percentage, as this would involve manually checking the results.

A training text of 361 test cases was constructed from the Hansards corpus, and a cross-validation text of 137 cases. The training was completed in 30 seconds and the trained network was used to label the sentence boundaries in a separate portion of the Hansards corpus containing 1546 punctuation marks with a lower bound of 83.0%. The SATZ system produced an accuracy of 99.0% on this text. Repeating the training and testing with a lower-case-only format also gave an accuracy rate of 99.0%.

# 7    Conclusions and Future Directions

The SATZ system offers a robust, rapidly trainable alternative to existing systems, which usually require extensive manual effort to develop and are specifically tailored to a text genre or natural language. By using prior probabilities of a word's part-of-speech to represent the context in which the word appears, the system offers significant savings in parameter estimation and training time. Although the systems of Wasson and Riley (1989) report slightly better error rates, the SATZ approach has the advantage of flexibility for application to new text genres, small training sets (and thereby fast training times), (relatively) small storage requirements, and little manual effort.

The boundary labeler was designed to be easily portable to new natural languages, assuming the accessibility of lexical part-of-speech frequency data (which can be obtained by running a part-of-speech tagger over a corpus of text, if it is not already available in the tagger itself). The success of applying SATZ to German and French with limited lexica and the experiments in English lexicon size described in Section 5.6 indicate that the lexicon itself need not be exhaustive. The heuristics used within the system to classify unknown words can compensate for inadequacies in the lexicon, and these heuristics can be easily adjusted to improve performance with a new language. I am currently working to adapt the SATZ system to additional languages, including Dutch, Italian, and Spanish. Since these languages are very similar to the three in which SATZ has already been implemented, it will be interesting to investigate its effectiveness with languages having different punctuation systems, such as Chinese or Arabic.

---

[22]The lexicon and all French texts (training, cross-validation, and test) were constructed by Marti Hearst at Xerox PARC.

While the results presented here indicate that the system in its current incarnation gives good results, many variations remain to be tested. It would be interesting to systematically investigate the effects of asymmetric context sizes, varied part-of-speech categorizations, abbreviation classes, and larger descriptor arrays. Although the neural network used in this system provides a simple, trainable tool for disambiguation, it would be instructive to compare its efficacy to a similar system which uses more conventional NLP tools such as Hidden Markov Models or decision trees.

In section 4.2.1 I discussed the representation of context and explained the processing circularity which makes it impossible to obtain the parts-of-speech from a tagger since the tagger requires the sentence boundaries. It would be possible to use the sentence boundaries labeled by SATZ to then use a tagger to obtain a single part-of-speech for each word. It would then be interesting to see if this more exact part-of-speech data would improve the accuracy of SATZ on the same text.

As discussed in section 1.1, there are several issues in sentence boundary disambiguation I have not addressed, but the methods developed in the SATZ system could be extended to such tasks. For example, different types of sentence boundaries, such as the "embedded end-of-sentence," could be identified in much the same manner, by including the necessary information in the training text and by adding output nodes to the neural network. Another potential extension of the SATZ system and a further test of its ability to adapt to new text types would be to train and run it on OCR'ed text, perhaps to assist in distinguishing punctuation marks or letters.

## Acknowledgements

This work would not have been possible without the assistance of Marti Hearst of Xerox PARC, who gave me invaluable guidance through each step of the process. I would also like to thank my research advisor, Prof. Robert Wilensky, and Prof. Jerome Feldman for reading drafts of this report and providing helpful suggestions for its improvement. Thanks also to the other members of the Berkeley Artificial Intelligence Research group for advice and technical support. In the course of this work, I was supported by a GAANN Fellowship and by the Advanced Research Projects Agency under Grant No. MDA972-92-J-1029 with the Corporation for National Research Initiatives (CNRI).

# Appendix A - README to accompany software version 1.0 of SATZ

This document gives information about the various parts of the SATZ
program.  Any questions should be directed at dpalmer@cs.berkeley.edu

```
FILES (full descriptions in files themselves):
getpart.c   looks up the token in the lexicon
lex.yy.c    tokenization (created by lex)
netinput.c    formats input to neural net
tagfile.c    labels sentence boundaries in input file
train.c    trains neural net
utilities.c    utilities used by above modules
common.h    all the #defines which can be altered for the program
trans.h    mapping of part-of-speech tags to descriptor array slots
utilities.h    function prototypes for utilities.c
weights.net    trained weights for neural net (created by train.c)
tokenize.l    tokenization file for use by lex
```

UNIX scripts: these scripts provide all the functionality for the
program.  File names within the scripts can be changed at
will, as long as you know what you are doing.  Just make sure
all the file names exist before running the script.

getfreqs: tokenizes file, looks up in lexicon (this script is good for
seeing if the lookup is working properly, but is not
really a part of SATZ itself
 usage: getfreqs {input file}

trainnet: trains neural net
usage: trainnet {file with training text}

bound: labels boundaries in input file
usage: bound {file to label}

Dictionary files: all dictionaries, or more accurately "word lists",

must contain lines in the following format in order to be readable:
word {tab} TAG1/freq1 {tab} TAG2/freq2 {tab} TAGn/freqn
example:  fixed\tJJ/1\tVBD/12\tVBN/69
(Note: {tab} is \t)
Important: do NOT leave a tab at the end of the line!

```
abbrev.dict     abbreviation dictionary
chars.dict      list of necessary characters or char strings (essential)
endings.dict    list of word endings (used to guess plurals, gerunds, etc.)
propnoun.dict   list of proper nouns (optional)
words.dict main lexicon
```

So, to use SATZ properly, you need to do the following:

1) Prepare a training text, which should be an excerpt from the text
to be labeled, of about 300-500 sentences.  All boundaries in
the training text must be labeled with the character sequence
</s>, or whatever sequence you use

2) Prepare a cross-validation text in the same manner.
It should be about 150-300 sentences, roughly half the size of
the training text.  Make sure the script "trainnet" has the
name of this text.

3) Run "trainnet" on the training text.  Training time varies, but
should be between 20 seconds and 5 minutes.  You should see
the progress of the training on the screen.  Net weights will
be stored in weights.net for use by "bound".  It is always
possible that the net won't behave nicely, as neural nets are
sometimes prone to do with the backprop algorithm.  If this
happens, you can try several things : modify the training text
and/or cross-validation text slightly, change the learning rate
(ETA in common.h) if the learning is oscillating
significantly, change BECOME_STABLE or STAY_STABLE in common.h
which determine length of training based on behavior of the
cross-validation text, or simply yell lots of obscenities at
the author.

4) Run "bound" on your files.  The output is sent to the file name
specified in the "bound" script, so you can change it
each time if you want.  Just be careful not to overwrite
previously labeled text if you label more than one file.

For each new language SATZ is used with, you will need to adjust a
few language-specific things, in order to maximize performance:

1) Make sure the tagset properly maps into the descriptor array.  This
is specified in the file "trans.h".  Four of the slots in
the descriptor array are reserved and must stay the same
regardless of the mapping (assuming 20 elements in array):
element 0 miscellaneous/other
element 17 first character capitalized
element 18 capital letter after possible sentence end
element 19 possible end-of-sentence punctuation mark

2) Change several #define declaration in common.h, including
DESPERATION_LABEL - a rough estimate of unknown word distribution
HYPHEN_LABEL - distribution for unknown hyphenated words
PROPER_NOUN_FACTOR -
PROPER_NOUN_AFTER_DOT

3) Make sure you have a lexicon of words in that language with prior
part-of-speech frequencies in the format above.  A list of
abbreviations and a list of proper nouns is optional.  I use
a very small (200 items) abbreviation list and no proper
noun list and get good results.  Just make sure the abbreviation
list includes the essential important ones (in English, Mr.,Dr.,
etc.) If you include these lists, make sure no members of the lists
are duplicated in two lists with conflicting pos labels.  Also, a
list of the most frequently encountered word endings in the
language will probably improve performance, as this list is used
to guess unknown words based on their endings (plurals, gerunds, etc.).
The endings file *must* contain at least one entry in order for the program to function properly.  (You can simply enter "zzzzzzz\tZ/1"
if you don't want this part to be used, or you could modify the
code to never access the endings file.)

4) The file chars.dict contains lots of standard characters encountered, so
you don't need to put them in the word-dictionary.  The token
**end** is returned by the tokenizer as a flag for training and
testing, whenever it encounters the string </s> in the text.
I chose this string assuming it would never occur naturally in any
text.  Feel free to change it, but make sure the string you choose
is included in the lexicon with the label ./2